# Consistent Multilabel Ranking through Univariate Loss Minimization

**Krzysztof Dembczyński**[1]      KDEMBCZYNSKI@CS.PUT.POZNAN.PL
**Wojciech Kotłowski**[1]      WKOTLOWSKI@CS.PUT.POZNAN.PL
**Eyke Hüllermeier**[2]      EYKE@INFORMATIK.UNI-MARBURG.DE

[1]Institute of Computing Science, Poznań University of Technology, Piotrowo 2, 60-965 Poznań, Poland
[2]Mathematics and Computer Science, Marburg University, Hans-Meerwein-Str., 35032 Marburg, Germany

## Abstract

We consider the problem of rank loss minimization in the setting of multilabel classification, which is usually tackled by means of convex surrogate losses defined on *pairs of labels*. Very recently, this approach was put into question by a negative result showing that commonly used pairwise surrogate losses, such as exponential and logistic losses, are inconsistent. In this paper, we show a positive result which is arguably surprising in light of the previous one: the simpler univariate variants of exponential and logistic surrogates (i.e., defined on *single labels*) are consistent for rank loss minimization. Instead of directly proving convergence, we give a much stronger result by deriving regret bounds and convergence rates. The proposed losses suggest efficient and scalable algorithms, which are tested experimentally.

## 1. Introduction

The problem of *multilabel classification* (MLC) has received increasing attention in machine learning research in recent years (Schapire & Singer, 2000; Elisseeff & Weston, 2001; Dekel et al., 2003; Dembczyński et al., 2010). In contrast to conventional (single-label) classification, where each instance is associated with a unique class label, MLC allows an instance to belong to several classes simultaneously. In other words, the "ground truth" is now a *subset* of positive labels instead of a single label. Correspondingly, more complex models need to be trained for predictive purposes, and their predictions need to be evaluated in terms of generalized loss functions.

Instead of producing predictions in terms of label subsets, one often prefers a *multilabel ranking*, that is, a ranking of labels from most likely positive to most likely negative. A prediction of that kind is commonly evaluated in terms of the *rank loss*, namely the fraction of incorrectly ordered label pairs; a positive and a negative label are incorrectly ordered if, in the predicted ranking, the former does not precede the latter—as it actually should do.

Many methods for MLC are based on the direct minimization of the number of conflicts, that is, pairwise ranking errors; more specifically, since the rank loss is highly discontinuous, such methods typically seek to minimize a convex surrogate. Interestingly, this approach has recently been called into question by Gao & Zhou (2011) (following the earlier results of Duchi et al. (2010)), who showed that the most commonly used convex surrogates of that kind are inconsistent.

In this paper, we complement this negative result by a positive one. More specifically, we prove that common convex surrogates used for binary classification, namely exponential and logistic losses, are consistent for the minimization of rank loss. Surprisingly, our surrogates are even simpler than existing ones for ranking, as they are *univariate* loss functions; thus, being defined on single labels rather than label pairs, it comes with additional advantages in terms of complexity. Instead of directly proving convergence, we give a much stronger result by deriving regret bounds and convergence rates.

The paper is organized as follows. In the next section, we introduce the setting of multilabel classification and elaborate on the rank loss for performance evaluation. Our main theoretical result is presented in Section 3 and discussed against the background of (Gao & Zhou, 2011) in Section 4. The theoretical contribution of the paper is complemented by some computational experiments in Section 5, prior to concluding with a summary in Section 6.

## 2. Multilabel Classification

In this section, we explain the MLC problem more formally and, along the way, introduce the notation used throughout the paper.

Let $\mathcal{X}$ denote an instance space, and let $\mathcal{L} = \{\lambda_1, \lambda_2, \ldots, \lambda_m\}$ be a finite set of class labels. We assume that an instance $x \in \mathcal{X}$ is (non-deterministically) associated with a subset of labels $L \in 2^{\mathcal{L}}$; this subset is often called the set of relevant (positive) labels, while the complement $\mathcal{L} \setminus L$ is considered as irrelevant (negative) for $x$. We identify a set $L$ of relevant labels with a binary vector $\boldsymbol{y} = (y_1, y_2, \ldots, y_m)$, in which $y_i = 1$ iff $\lambda_i \in L$. The set of possible labelings is denoted $\mathcal{Y} = \{0,1\}^m$. We assume observations to be generated independently and randomly according to a probability distribution $P(X = x, \boldsymbol{Y} = \boldsymbol{y})$ (later denoted $P(x, \boldsymbol{y})$) on $\mathcal{X} \times \mathcal{Y}$, i.e., an observation $\boldsymbol{y} = (y_1, \ldots, y_m)$ is the realization of a corresponding random vector $\boldsymbol{Y} = (Y_1, Y_2, \ldots, Y_m)$.

A multilabel classifier $\boldsymbol{h}$ assigns a (predicted) label subset to each instance $x \in \mathcal{X}$. More generally, we allow the output of the classifier to be a vector of real numbers $\boldsymbol{h}(x) = (h_1(x), \ldots, h_m(x)) \in \mathbb{R}^m$, which means that $\boldsymbol{h}$ is an $\mathcal{X} \to \mathbb{R}^m$ mapping. A score vector of this kind can not only be turned into a label subset (binary vector $\boldsymbol{y} \in \{0,1\}^m$) via thresholding, but can also be used for ranking the labels $\lambda_i$ in a natural way, namely by sorting them in decreasing order according to their respective scores $s_i = h_i(x)$.

### 2.1. Loss, Risk and Regret

The prediction accuracy of $\boldsymbol{h}$ is measured in terms of its *risk*, that is, its expected (classification) loss

$$L(\boldsymbol{h}, P) = \mathbb{E}\left[\ell(\boldsymbol{Y}, \boldsymbol{h}(X))\right] = \int \ell(\boldsymbol{y}, \boldsymbol{h}(x)) \, dP(x, \boldsymbol{y}), \quad (1)$$

where $\ell \colon \mathcal{Y} \times \mathbb{R}^m \to \mathbb{R}$ is a *loss function*. In addition, it will be convenient to use an expected loss conditioned on an instance $x \in \mathcal{X}$:

$$L(\boldsymbol{h}, P \,|\, x) = \mathbb{E}\left[\ell(\boldsymbol{Y}, \boldsymbol{h}(x)) \,|\, x\right] = \sum_{\boldsymbol{y} \in \mathcal{Y}} \ell(\boldsymbol{y}, \boldsymbol{h}(x)) P(\boldsymbol{y} \,|\, x),$$

so that $L(\boldsymbol{h}, P) = \mathbb{E}[L(\boldsymbol{h}, P \,|\, X)]$.

The risk of a classifier is not always a good indicator of its true performance, as it does not account for the hardness of the problem. In fact, even the optimal classifier $\boldsymbol{h}^*$ (which has access to the distribution $P(x, \boldsymbol{y})$) will normally have a non-zero risk. We call $\boldsymbol{h}^*$ the *Bayes classifier*. For each $x \in \mathcal{X}$, this classifier minimizes expected loss conditioned on $x$:

$$\boldsymbol{h}^*(x) = \arg\min_{\boldsymbol{s} \in \mathbb{R}^m} \sum_{\boldsymbol{y} \in \mathcal{Y}} \ell(\boldsymbol{y}, \boldsymbol{s}) P(\boldsymbol{y} \,|\, x) \quad (2)$$

We note that in general, $\boldsymbol{h}^*$ is not unique. However, the risk of $\boldsymbol{h}^*$, denoted $L^*(P)$, is unique, and is called the *Bayes risk*. It offers a reasonable baseline for comparison and suggests to define the *regret* of a classifier $\boldsymbol{h}$ as follows:

$$\text{Reg}(\boldsymbol{h}, P) = L(\boldsymbol{h}, P) - L^*(P) \quad (3)$$

Occasionally, we will also use the regret conditioned on an instance $x$, denoted $\text{Reg}(\boldsymbol{h}, P \,|\, x)$. Later on, when analyzing the risk and regret for particular loss functions, such as rank loss, we will use more specific notations like $L_{\text{rnk}}$ and $\text{Reg}_{\text{rnk}}$, which indicate the loss function that risk and regret are referring to.

### 2.2. Rank Loss

In this paper, we focus on the *rank loss*, which is among the most important loss functions in MLC and has attracted much attention in recent years (Dembczyński et al., 2010; Gao & Zhou, 2011):

$$\ell_{\text{rnk}}(\boldsymbol{y}, \boldsymbol{h}) = w(\boldsymbol{y}) \sum_{(i,j) : y_i > y_j} \left( [\![h_i < h_j]\!] + \frac{1}{2}[\![h_i = h_j]\!] \right), \quad (4)$$

where $[\![\cdot]\!]$ is the standard {false, true} $\to \{0,1\}$ mapping (for the sake of clarity, we will suppress dependence on $x$ in the notation, whenever it is clear from the context). Treating the classifier's output as a ranking, the rank loss compares the true label subset with this ranking, in which all relevant labels ideally precede all irrelevant ones. More specifically, the rank loss counts the number of label pairs violating this condition and multiplies it by a positive weight $w(\boldsymbol{y})$. In other words, the "penalty" or "cost" for a mistake on a label pair is given by $w(\boldsymbol{y})$ and may thus depend on properties of the true labeling $\boldsymbol{y}$.

Typically, $w(\boldsymbol{y})$ is a normalization constant equal to the reciprocal of the total number of pairwise comparisons between labels, thus accounting for the fact that the maximum number of possible mistakes depends on the number of positive labels in $\boldsymbol{y}$. Yet, we shall not make any specific assumptions about about $w(\boldsymbol{y})$ throughout the paper, except that it is non-negative and bounded: $0 \leq w(\boldsymbol{y}) \leq w_{\max}$ for all $\boldsymbol{y}$.[1]

---

[1] For our results, it is even enough to assume that $w$ is bounded in expectation: $\mathbb{E}[w(\boldsymbol{Y}) \,|\, x] \leq w_{\max}$ for all $x$.

Let us determine the Bayes classifier for the rank loss. To this end, it is convenient to introduce the following quantity:

$$\Delta_{ij}^{uv} = \sum_{\boldsymbol{y}:\, y_i=u, y_j=v} w(\boldsymbol{y}) P(\boldsymbol{y} \,|\, x)\,,$$

where $i, j \in \{1, \ldots, m\}$ and $u, v \in \{0, 1\}$. Note that $\Delta_{ij}^{uv}$ reduces to the marginal probability $P(Y_i = u, Y_j = v \,|\, x)$ if $w(\boldsymbol{y}) \equiv 1$. For a more general weight function $w(\cdot)$, $\Delta_{ij}^{uv}$ combines the probability of the label combination $(Y_i = u, Y_j = v)$ with the potential penalty in case these labels are ranked incorrectly. Thus, it can be seen as a kind of importance of this label combination.

By definition, $\Delta_{ij}^{uv} = \Delta_{ji}^{vu}$ for all $(i, j)$ and

$$\Delta_{ij}^{00} + \Delta_{ij}^{01} + \Delta_{ij}^{10} + \Delta_{ij}^{11} = W\,,$$

where $W = \mathbb{E}[w(\boldsymbol{Y}) \,|\, x] = \sum_{\boldsymbol{y}} w(\boldsymbol{y}) P(\boldsymbol{y} \,|\, x)$ (which is a condition similar to the normalization property of a probability distribution). Then, the conditional risk can be written as follows:

$$L_{\text{rnk}}(\boldsymbol{h}, P \,|\, x) = \sum_{i>j} \Big( \Delta_{ij}^{10} [\![h_i > h_j]\!] + \Delta_{ij}^{01} [\![h_i < h_j]\!] \\ + \frac{1}{2}(\Delta_{ij}^{10} + \Delta_{ij}^{01}) [\![h_i = h_j]\!] \Big) \quad (5)$$

To proceed further, we define $\Delta_i^u = \Delta_{ij}^{u0} + \Delta_{ij}^{u1}$ for any $j \neq i$ (one readily verifies that this quantity does not depend on $j$). $\Delta_i^u$ plays a role comparable to the marginal probability $P(Y_i = u \,|\, x)$. We have $\Delta_i^0 + \Delta_i^1 = W$ for all $i$ and

$$\Delta_i^1 - \Delta_j^1 = \Delta_{ij}^{10} + \Delta_{ij}^{11} - \Delta_{ji}^{10} - \Delta_{ji}^{11} \qquad (6) \\ = \Delta_{ij}^{10} + \Delta_{ij}^{11} - \Delta_{ij}^{01} - \Delta_{ij}^{11} = \Delta_{ij}^{10} - \Delta_{ij}^{01}.$$

The Bayes classifier ranks labels according to the $\Delta_i^1$, i.e., a vector $\boldsymbol{h}^* = (h_1^*, \ldots, h_m^*)$ is a Bayes prediction if $h_i^* > h_j^*$ whenever $\Delta_i^1 > \Delta_j^1$, $h_i^* = h_j^*$ whenever $\Delta_i^1 = \Delta_j^1$, and $h_i^* < h_j^*$ whenever $\Delta_i^1 < \Delta_j^1$. Indeed, using (6), we see that the Bayes classifier thus defined minimizes every term in the sum in (5). This result extends the result by Gao & Zhou (2011) defined in terms of $\Delta_{ij}^{10}$. The Bayes risk conditioned on $x$ is given by

$$L_{\text{rnk}}^*(P \,|\, x) = \sum_{1 \leq i < j \leq m} \min\{\Delta_{ij}^{10}, \Delta_{ij}^{01}\}\,. \qquad (7)$$

The equality $\Delta_i^1 - \Delta_j^1 = \Delta_{ij}^{10} - \Delta_{ij}^{01}$ in (6) is not only useful but also remarkable. In order to understand its meaning, it is convenient to consider the special case $w(\boldsymbol{y}) \equiv 1$, in which the $\Delta$-values reduce to conditional probabilities (Dembczyński et al., 2010). In this case, (6) becomes

$$P(Y_i = 1, Y_j = 0 \,|\, x) - P(Y_i = 0, Y_j = 1 \,|\, x) \\ = P(Y_i = 1 \,|\, x) - P(Y_j = 1 \,|\, x)\,.$$

The decision whether label $\lambda_i$ should be ranked ahead of $\lambda_j$ or the other way around depends on the sign of the left-hand side: If the *joint* probability of $(Y_i = 1, Y_j = 0)$ is higher than the *joint* probability of $(Y_i = 0, Y_j = 1)$, the answer should be affirmative, otherwise not. According to the above equation, the answer can be found by just looking at the *marginal* probabilities $P(Y_i = 1 \,|\, x)$ and $P(Y_j = 1 \,|\, x)$. This is remarkable, as it means that the dependency between $Y_i$ and $Y_j$ can safely be ignored—a key observation for our main result in the next section.

## 3. Main Result

We prepare our main result, to be presented in Section 3.3, by two auxiliary results.

First, in Section 3.1, we show that rank regret depends solely on the marginal weights $\Delta_i^u$, and that we are allowed to replace the original distribution $P$ by any other distribution $P'$, as long as they both lead to the same marginal weights $\Delta_i^u$. In particular, we can choose $P'$, for which labels are (conditionally) independent.

Second, in Section 3.2, we provide the basic argument for the use of univariate loss functions, showing that, under the assumption of independence, such losses are sufficient for the consistent ranking of objects. This result will be shown, not for MLC directly, but in the context of the related problem of bipartite ranking.

The final step in Section 3.3 will therefore consist of transferring this result back to the setting of MLC, using the trick from Section 3.1 and the fact that expected univariate losses depend on distribution only through the marginal weights $\Delta_i^u$.

### 3.1. Label Dependence Does Not Influence Rank Regret

The main problem in the analysis of the regret (3) is the conditional dependence of labels given $x$. As already mentioned, however, this dependence does not seem to play an important role in the minimization of rank regret. In the following, we shall make this observation more explicit by showing that rank regret depends solely on the marginal weights $\Delta_i^u$:

**Lemma 3.1.** *For every $x \in \mathcal{X}$, and every multilabel*

*classifier* $\boldsymbol{h}$:

$$\text{Reg}(\boldsymbol{h}, P \,|\, x) = \sum_{1 \leq j < i \leq m} \left( \Delta_i^1 [\![h_i < h_j]\!] + \Delta_j^1 [\![h_i > h_j]\!] \right.$$
$$\left. + \frac{\Delta_i^1 + \Delta_j^1}{2} [\![h_i = h_j]\!] - \min\{\Delta_i^1, \Delta_j^1\} \right).$$

*Proof.* According to (5) and (7), $\text{Reg}(\boldsymbol{h}, P \,|\, x)$ can be written as

$$\text{Reg}(\boldsymbol{h}, P \,|\, x) = \sum_{1 \leq j < i \leq m} B_{ij},$$

where

$$B_{ij} = \Delta_{ij}^{10} [\![h_i < h_j]\!] + \Delta_{ij}^{01} [\![h_i > h_j]\!]$$
$$+ \frac{\Delta_{ij}^{10} + \Delta_{ij}^{01}}{2} [\![h_i = h_j]\!] - \min\{\Delta_{ij}^{10}, \Delta_{ij}^{01}\}.$$

Since only one of the first three terms can be nonzero, $B_{ij}$ will not change if we add $\Delta_{ij}^{11}$ to the first three terms and subtract it from the last term:

$$B_{ij} = (\Delta_{ij}^{10} + \Delta_{ij}^{11})[\![h_i < h_j]\!] + (\Delta_{ij}^{01} + \Delta_{ij}^{11})[\![h_i > h_j]\!]$$
$$+ \frac{(\Delta_{ij}^{10} + \Delta_{ij}^{11}) + (\Delta_{ij}^{01} + \Delta_{ij}^{11})}{2}[\![h_i = h_j]\!]$$
$$- \min\{\Delta_{ij}^{10} + \Delta_{ij}^{11}, \Delta_{ij}^{01} + \Delta_{ij}^{11}\}.$$

By definition, $\Delta_{ij}^{10} + \Delta_{ij}^{11} = \Delta_i^1$, $\Delta_{ij}^{01} + \Delta_{ij}^{11} = \Delta_{ji}^{10} + \Delta_{ji}^{11} = \Delta_j^1$, so that:

$$B_{ij} = \Delta_i^1 [\![h_i < h_j]\!] + \Delta_j^1 [\![h_i > h_j]\!]$$
$$+ \frac{\Delta_i^1 + \Delta_j^1}{2} [\![h_i = h_j]\!] - \min\{\Delta_i^1, \Delta_j^1\}. \quad (8)$$

$\square$

Lemma 3.1 implies that the rank regret of any multilabel classifier $\boldsymbol{h}$ will not change if we replace the original distribution $P$ and weight function $w$ by any other distribution $P'$ and function $w'$, as long as they lead to the same marginal weights $\Delta_i^u$. In particular, we can choose $P'$ to be a product distribution, for which labels are (conditionally) independent, and the constant weight function $w'(\boldsymbol{y}) = W$ for all $\boldsymbol{y}$. As we shall see in Section 3.3, this will effectively result in a bipartite ranking problem for every $x$.

Before exploiting this finding in Section 3.3, we provide a second building block of our main result, showing that the minimization of specific univariate losses is sufficient for the proper ranking of objects under the assumption of independence. To this end, we refer to the related though slightly simpler setting of bipartite ranking. From now on, it will be more convenient to encode labels as $-1$ and $+1$, i.e., $y_i \in \{-1, +1\}$ instead of $\{0, 1\}$.

### 3.2. Univariate Loss Minimization is Sufficient under the Assumption of Independence

The *bipartite ranking* problem (Cohen et al., 1999; Clémençon et al., 2008; Kotłowski et al., 2011) is in a sense in-between MLC and standard binary classification. Like in the latter, there is only a single binary class label, but like in MLC, performance is measured in terms of rank loss instead of classification error. However, instead of ranking labels given an instance, the problem is to rank the instances themselves.

More specifically, consider a simple binary classification problem with training examples $(\tilde{x}, \tilde{y}) \in \tilde{\mathcal{X}} \times \{-1, +1\}$. A classifier $\tilde{h}$ is a real-valued function $\tilde{h} \colon \tilde{\mathcal{X}} \to \mathbb{R}$, and performance is measured in terms of a *bipartite rank loss* defined on pairs of labels:

$$\ell_{\text{br}}((\tilde{y}, \tilde{y}'), (\tilde{h}, \tilde{h}')) = [\![\tilde{y} > \tilde{y}']\!] [\![\tilde{h} < \tilde{h}']\!]$$
$$+ [\![\tilde{y} < \tilde{y}']\!] [\![\tilde{h} > \tilde{h}']\!] + \frac{1}{2}[\![\tilde{y} \neq \tilde{y}']\!] [\![\tilde{h} = \tilde{h}']\!]$$

This is a *non-normalized* version of the rank loss, which is more useful for our purposes; in the literature, it is common to use a normalized version, which differs from the non-normalized one by a product of class priors (Clémençon et al., 2008).

Given the loss, we can define risk and regret by taking expectations over the pairs of instances which are generated i.i.d.:

$$L_{\text{br}}(\tilde{h}, \tilde{P}) = \mathbb{E}[\ell_{\text{br}}((\tilde{Y}, \tilde{Y}'), (\tilde{h}(\tilde{X}), \tilde{h}(\tilde{X}')))]$$
$$= \int \ell_{\text{br}}((\tilde{y}, \tilde{y}'), (\tilde{h}(\tilde{x}), \tilde{h}(\tilde{x}'))) d\tilde{P}(\tilde{x}, \tilde{y}) d\tilde{P}(\tilde{x}', \tilde{y}'),$$
$$\text{Reg}_{\text{br}}(\tilde{h}, \tilde{P}) = L_{\text{br}}(\tilde{h}, \tilde{P}) - \inf_{\tilde{h}'} L_{\text{br}}(\tilde{h}', \tilde{P}).$$

Let $\ell_{\exp}(\tilde{y}, \tilde{h}) = e^{-\tilde{y}\tilde{h}}$, $\ell_{\log}(\tilde{y}, \tilde{h}) = \log(1 + e^{-\tilde{y}\tilde{h}})$ be the standard exponential and logistic losses for binary classification. For these losses, we can again define risks $L_{\exp}(\tilde{h}, \tilde{P}), L_{\log}(\tilde{h}, \tilde{P})$ and regrets $\text{Reg}_{\exp}(\tilde{h}, \tilde{P}), \text{Reg}_{\log}(\tilde{h}, \tilde{P})$ in a standard way. The following theorem relates bipartite ranking regret to regrets in terms of exponential and logistic loss:

**Theorem 3.1.**

$$\text{Reg}_{\text{br}}(\tilde{h}, \tilde{P}) \leq \sqrt{\frac{3}{2}} \sqrt{\text{Reg}_{\exp}(\tilde{h}, \tilde{P})} \quad (9)$$

$$\text{Reg}_{\text{br}}(\tilde{h}, \tilde{P}) \leq \sqrt{2} \sqrt{\text{Reg}_{\log}(\tilde{h}, \tilde{P})} \quad (10)$$

Theorem 3.1 is very similar to Theorem 4.1 in (Kotłowski et al., 2011), except that the latter involves a normalized version of the bipartite rank loss and so-called balanced loss functions. Nevertheless, in order to show Theorem 3.1, the proof from (Kotłowski et al., 2011) can be adapted quite easily. Here, we omit a detailed presentation of the modifications required due to space restrictions.

### 3.3. Minimizing Rank Loss in MLC

The exponential loss and the logistic loss introduced above are commonly used in standard classification. A straightforward extension of these losses to the MLC setting, taking multiple labels and instance weights into account, is given as follows:

$$\ell_{\exp}(\boldsymbol{y}, \boldsymbol{h}) = w(\boldsymbol{y}) \sum_{i=1}^{m} e^{-y_i h_i}, \quad (11)$$

$$\ell_{\log}(\boldsymbol{y}, \boldsymbol{h}) = w(\boldsymbol{y}) \sum_{i=1}^{m} \log\left(1 + e^{-y_i h_i}\right). \quad (12)$$

The minimization of these losses comes down to solving $m$ independent classification problems, one for each label. Any algorithm for classification with exponential or logistic surrogate, such as AdaBoost or logistic regression, can be used for this purpose, provided it allows for handling weighted instances. Despite its simplicity and efficiency, this approach provides a consistent way of minimizing the rank loss, as shown by the following result.

**Theorem 3.2.** *Let* $\text{Reg}_{\exp}(\boldsymbol{h}, P)$ *and* $\text{Reg}_{\log}(\boldsymbol{h}, P)$ *be the regrets for exponential and logistic losses, respectively. Then*

$$\text{Reg}_{\text{rnk}}(\boldsymbol{h}, P) \leq \frac{\sqrt{6}}{4} C \sqrt{\text{Reg}_{\exp}(\boldsymbol{h}, P)}, \quad (13)$$

$$\text{Reg}_{\text{rnk}}(\boldsymbol{h}, P) \leq \frac{\sqrt{2}}{2} C \sqrt{\text{Reg}_{\log}(\boldsymbol{h}, P)}, \quad (14)$$

*where* $C \leq m\sqrt{m w_{\max}}$.

*Proof.* The idea of the proof is to reduce an MLC problem, conditioned on an instance $x$, to a bipartite ranking problem, which then allows us to exploit Theorem 3.1. More specifically, for a given $x$, we define a bipartite ranking problem by setting $\tilde{\mathcal{X}} = \{1, \ldots, m\}$; that is, the objects (instances) to be ranked now correspond to the label indices of our MLC problem and are of the form $\tilde{x} = i$, $(i = 1, \ldots, m)$. Moreover, we define a distribution $\tilde{P}$ on $\tilde{\mathcal{X}} \times \{-1, +1\}$ as follows:

$$\tilde{P}(\tilde{X} = i) = \frac{1}{m}, \quad \tilde{P}(\tilde{Y} = 1 \mid \tilde{X} = i) = \frac{\Delta_i^1}{W} \quad (15)$$

For a classifier $\tilde{h}$ with $\tilde{h}(\tilde{x} = i) = h_i$, it is easy to see that

$$\text{Reg}_{\text{br}}(\tilde{h}, \tilde{P}) = \frac{1}{m^2} \sum_{i,j} \tilde{B}_{ij},$$

where $\tilde{B}_{ij}$ is defined as:

$$\tilde{B}_{ij} = \frac{\Delta_i^1}{W}\left(1 - \frac{\Delta_j^1}{W}\right)[\![h_i < h_j]\!] + \frac{\Delta_j^1}{W}\left(1 - \frac{\Delta_i^1}{W}\right)[\![h_i > h_j]\!]$$

$$+ \frac{1}{2}\left(\frac{\Delta_i^1}{W}\left(1 - \frac{\Delta_j^1}{W}\right) + \frac{\Delta_j^1}{W}\left(1 - \frac{\Delta_i^1}{W}\right)\right)[\![h_i = h_j]\!]$$

$$- \min\left\{\frac{\Delta_i^1}{W}\left(1 - \frac{\Delta_j^1}{W}\right), \frac{\Delta_j^1}{W}\left(1 - \frac{\Delta_i^1}{W}\right)\right\} = \frac{B_{ij}}{W},$$

where the last equality is valid because the term $\frac{\Delta_i^1}{W}\frac{\Delta_j^1}{W}$ cancels and because of (8). Using the above and Lemma 3.1, we see that

$$\text{Reg}_{\text{br}}(\tilde{h}, \tilde{P}) = \frac{1}{m^2}\sum_{i,j}\tilde{B}_{ij} = \frac{1}{Wm^2}\sum_{i,j}B_{ij}$$

$$\geq \frac{2}{Wm^2}\sum_{1 \leq j < i \leq m} B_{ij} = \frac{2}{Wm^2}\text{Reg}_{\text{rnk}}(\boldsymbol{h}, P|x). \quad (16)$$

Theorem 3.1 relates $\text{Reg}_{\text{br}}(\tilde{h}, \tilde{P})$ to $\text{Reg}_\ell(\tilde{h}, \tilde{P})$, for $\ell$ being the exponential or logistic loss. What remains, therefore, is to trace back $\text{Reg}_\ell(\tilde{h}, \tilde{P})$ to $\text{Reg}_\ell(\boldsymbol{h}, P \mid x)$, where the latter regret is based on the original distribution $P$ and the multilabel versions (11–12) of exponential and logistic loss. The following equalities hold:

$$L_\ell(\boldsymbol{h}, P \mid x) = \sum_{i=1}^{n} \ell(1, h_i)\Delta_i^1 + \ell(-1, h_i)\Delta_i^0,$$

$$L_\ell^*(P \mid x) = \sum_{i=1}^{n} \inf_h \left\{\ell(1, h)\Delta_i^1 + \ell(-1, h)\Delta_i^0\right\},$$

where we have, respectively, risks based on the multilabel loss and risks based on standard classification loss on the left-hand and right-hand side. Due to (15), we get

$$\text{Reg}_\ell(\boldsymbol{h}, P \mid x) = Wm\text{Reg}_\ell(\tilde{h}, \tilde{P}). \quad (17)$$

Taking (16), (9–10), and (17) together gives

$$\text{Reg}_{\text{rnk}}(\boldsymbol{h}, P \mid x) \leq \frac{\sqrt{6}}{4} C\sqrt{\text{Reg}_{\exp}(\boldsymbol{h}, P \mid x)},$$

$$\text{Reg}_{\text{rnk}}(\boldsymbol{h}, P \mid x) \leq \frac{\sqrt{2}}{2} C\sqrt{\text{Reg}_{\log}(\boldsymbol{h}, P \mid x)},$$

where $C = m\sqrt{mW}$. The Theorem is proved by noting that $W \leq w_{\max}$, taking the expectation with respect to $x$ on both sides, and applying Jensen inequality $\mathbb{E}[f(X)] \leq f(\mathbb{E}[X])$ for the concave function $f(x) = \sqrt{x}$. □

One might be concerned by the possibly large constant $C = m\sqrt{mW}$ appearing in the bound, wondering whether it could perhaps be improved. However, $C$ is indeed *expected* to appear in the bound and does actually not weaken it. Instead, it only compensates for the difference in the scale of both sides of (13) and (14). Indeed, the rank regret on the left-hand side scales like $O(m^2 W)$, while the square root of exponential/logistic regret on the right-hand side scales like $O(\sqrt{mW})$. Therefore, there *must* be a constant $O(m\sqrt{mW})$ on the right-hand side to compensate for the difference.

Another question is whether the square-root convergence in (13–14) could be improved. The answer is negative: Bartlett et al. (2006) already showed for binary classification (which can be casted as a special MLC ranking problem) that the square-root bound is unavoidable in the worst case.

## 4. Relationship to Prior Work

This section is meant to look at the result of Gao & Zhou (2011) against the background of our findings so far, trying to support a more intuitive understanding. As mentioned earlier, these authors consider pairwise convex surrogate losses of the form

$$\ell_\phi(\boldsymbol{y}, \boldsymbol{h}) = \sum_{(i,j):\, y_i > y_j} w(\boldsymbol{y}) \phi(h_i - h_j), \quad (18)$$

where $\phi$ is a convex, differential, non-linear, and non-increasing function, and show that no such loss is consistent for multilabel ranking. Given the existence of pairwise losses that are actually consistent for *bipartite ranking*, this result appears to be surprising at first sight, all the more since, in our proof, we are using a reduction to bipartite ranking, too.

The reason for inconsistency becomes more apparent when looking at the conditional expected loss:

$$L_\phi(\boldsymbol{h}, P \,|\, x) = \sum_{i>j} \Delta_{ij}^{10} \phi(h_i - h_j) + \Delta_{ji}^{10} \phi(h_j - h_i) \quad (19)$$

A necessary condition for consistency is that the Bayes classifier $\boldsymbol{h}^*$ for $\phi$-loss is also the Bayes ranker, i.e.,

$$\operatorname{sign}(h_i^* - h_j^*) = \operatorname{sign}(\Delta_{ij}^{10} - \Delta_{ij}^{01}). \quad (20)$$

To ease understanding, it is again convenient to consider the special case $w(\boldsymbol{y}) \equiv 1$, in which the $\Delta$-values reduce to conditional probabilities (and, therefore, are more easily interpretable). In our approach of univariate loss minimization, (20) is indeed valid: According to (6), the equality $\Delta_i^1 - \Delta_j^1 = \Delta_{ij}^{10} - \Delta_{ij}^{01}$ holds true. Moreover, by applying a convex loss function $\phi$, the prediction $h_i^*$ of the Bayes classifier is a nonlinear yet *monotone* transformation of the conditional probability $\Delta_i^1 = P(Y_i = 1 \,|\, x)$: The larger the probability of the conditional class, the larger the score produced by the Bayes classifier. Consequently, $\operatorname{sign}(h_i^* - h_j^*) = \operatorname{sign}(\Delta_i^1 - \Delta_j^1) = \operatorname{sign}(\Delta_{ij}^{10} - \Delta_{ij}^{01})$. Thus, loosely speaking, our approach guarantees consistency because the (Bayes) decision of how to rank two labels $\lambda_i$ and $\lambda_j$, which depends on the sign of $\Delta_{ij}^{10} - \Delta_{ij}^{01}$, remains unaffected by both of our measures: the consideration of univariate marginals $\Delta_i^1$ instead of the bivariate ones $\Delta_{ij}^{10}$ and $\Delta_{ij}^{01}$ (label dependency does not matter), as well as the transformation implied by the convex surrogate afterward.

Now, although the first argument of the irrelevance of label dependence does in principle remain valid in case of pairwise loss, consistency is essentially lost in the second step. In fact, the use of a convex surrogate loss has a much more involved effect in the pairwise case, since the (nonlinear monotone) transformation now applies to the *differences* $\Delta_i^1 - \Delta_j^1 = P(Y_i = 1 \,|\, x) - P(Y_j = 1 \,|\, x)$ of conditional probabilities instead of the conditionals themselves. Therefore, since each $\Delta_i^1$ simultaneously participates in several such differences, the minimization of (19) results in a complicated solution $\boldsymbol{h}^*$, where $h_i^*$ generally depends on all $\Delta_{jk}^{10}$ ($1 \leq j, k \leq m$), and not only on $\Delta_i^1$. Gao & Zhou (2011) exploit this observation to show that, for any $\phi$ defined as above, the pairwise marginals $\Delta_{jk}^{10}$ can be chosen such that the Bayes classifier $\boldsymbol{h}^*$ is *not* the Bayes ranker. The only case in which (19) admits a simple solution for certain losses $\phi$ is when the labels are independent, and this is exactly the case of bipartite ranking, for which consistency is known to hold.

We note that Gao & Zhou (2011) (following Duchi et al. (2010)) showed consistency (but not regret bounds and convergence rates) of some specific pairwise surrogate losses, one of which can be rewritten as a univariate *linear* surrogate with regularization.

## 5. Empirical Results

To verify our theoretical claims we performed experimental studies on synthetic and benchmark data. We measured the performance of the algorithms in terms of the rank loss (4) with weights defined as:

$$w(\boldsymbol{y}) = (s_{\boldsymbol{y}}(m - s_{\boldsymbol{y}}))^{-1}, \quad \text{where } s_{\boldsymbol{y}} = \sum_i y_i. \quad (21)$$

This is a popular choice, as the weights are the inverses of the total number of pairwise comparisons between labels. Thus, the value of the rank loss is between 0 (perfect ordering) and 1 (reversed ordering).

The main goal of the experiment is to verify whether simple algorithms based on univariate surrogate losses (11) and (12) are competitive to state-of-the-art al-gorithms that minimize the rank loss using convex pairwise surrogates (18). Note that minimization of (11) and (12) reduces to solving $m$ independent classification tasks with weighted training examples. In other words, each task is solved by using an algorithm that minimizes the ordinary exponential or logistic loss on a set of weighted training examples. We used AdaBoost.M1 to minimize the exponential loss and logistic regression to minimize the logistic loss. We refer to this reduction framework as Weighted Binary Relevance (WBR). We compared WBR with two well-known algorithms for multilabel ranking, AdaBoost.MR (Schapire & Singer, 2000) and log-linear models for label ranking (LLLR) (Dekel et al., 2003). These two algorithms seek to minimize the rank loss by using convex surrogates defined on label pairs (18). AdaBoost.MR uses the exponential and LLLR the logistic surrogate. Let us underline that both AdaBoost.MR and LLLR use weights (21) in their surrogates, so that all the algorithms are tailored for the same performance measure.

In boosting algorithms, we used decision stumps as weak learners. For AdaBoost.M1, we selected the number of decision stumps from $\{10, 20, 50, 100, 200\}$, while for AdaBoost.MR, the total number of decision stumps from $\{10, 20, 50, 10^2, \ldots, 10^4, 2 \times 10^4\}$. The regularization parameter in logistic regression was tuned in the range $\{10^{-3}, 10^{-2}, \ldots, 10^3\}$. We ran LLLR with different numbers of iterations $\{10, 20, 50, 10^2, \ldots, 10^4, 2 \times 10^4\}$.[2] The tuning process should not favor any of the methods, as all algorithms have a single parameter to choose.

### 5.1. Synthetic Data

We designed synthetic data to show a difference in the performance of the algorithms in the cases of label dependence and independence, respectively. The model is based on latent variables $\boldsymbol{f} = (f_1, f_2, \ldots, f_m)$:

$$\boldsymbol{f} = \boldsymbol{A}\boldsymbol{x} + \boldsymbol{\epsilon},$$

where $\boldsymbol{x}$ is a two-dimensional feature vector uniformly drawn from a unit disk, $\boldsymbol{A}$ is an $m \times 2$ matrix of linear coefficients, and $\boldsymbol{\epsilon}$ is an $m$-dimensional noise vector whose coordinates are drawn from $N(0, 0.25)$. The labels are obtained from the latent variables according to

$$\boldsymbol{y} = [\![ \boldsymbol{M}\boldsymbol{f} > \boldsymbol{0} ]\!],$$

where $\boldsymbol{M}$ is an $m \times m$ mixing matrix introducing dependencies between labels, and $[\![\cdot]\!]$ applies to each element of the vector separately. A single model is thus determined by the choice of $\boldsymbol{A}$ and $\boldsymbol{M}$. The independent label case is obtained for $\boldsymbol{M}$ being the identity $\boldsymbol{I}$.

We generated 10 random models, with rows of $\boldsymbol{A}$ drawn uniformly from a 2-dimensional unit sphere. For each model, we considered the case of independent ($\boldsymbol{M} = \boldsymbol{I}$) and dependent labels (entries of $\boldsymbol{M}$ drawn independently and uniformly from $[-1, 1]$). In each case, we trained all 4 algorithms on training sets of different sizes $n$, varying $n$ from 100 to 16000 examples. For each $n$, 10 training sets of a given size were generated (thus, there are 100 repetitions for any given training set size $n$). For testing, we used a dataset containing 50 000 examples.

The results are given in Fig. 1. We compared the algorithms based on exponential loss separately from those for logistic loss. The results shown in Fig. 1 are nicely in agreement with what we expect from our theoretical results, at least for the logistic loss: In the case of label independence, where both pairwise and univariate loss minimization are consistent, the methods perform more or less en par. However, in the case where labels are not independent, and hence the pairwise approach is no longer consistent, our approach of univariate loss minimization shows small but consistent improvements. In the case of exponential loss, the picture is not entirely clear,[3] but the univariate approach seems to outperform its competitor based on pairwise loss for large enough training data. Weaker performance of the exponential loss follows from the fact that the stumps used as base learners in boosting do not exactly match the true (linear) model.

### 5.2. Benchmark Data

We also performed an experiment on commonly used benchmark datasets.[4] We chose 4 datasets of moderate size, with around 10 labels, and one large dataset with 101 labels and more than 30K training examples. The datasets are described in Table 1. To facilitate comparison of the results presented in this paper, we used the original split for training and test sets.

The results are given in Table 2, again separately for exponential and logistic loss. The picture conveyed by these results is less clear than it was for the synthetic datasets. In fact, since the true nature of the data is not known (i.e., whether or not the labels are independent), is is difficult to draw clear conclusions. Nevertheless, one can safely say the simple reduction algorithms trained independently on each label are competitive to state-of-the-art algorithms defined on pairwise surrogates. Again, this is in complete agreement with our theoretical results.

---

[2] To perform experiments, we used MULAN (http://mulan.sourceforge.net/) and Weka (http://www.cs.waikato.ac.nz/ml/weka/) packages, implementation of logistic regression from Mallet (http://mallet.cs.umass.edu/), and the original implementation of AdaBoost.MR from the BoosTexter package (http://www.cs.princeton.edu/~schapire/boostexter.html).

[3] AdaBoost.MR behaves quite strangely on these datasets as for more than 20 stumps it quickly overfits. Therefore, we also limited the number of stumps in WBR-AdaBoost.

[4] These datasets are taken from MULAN http://mulan.sourceforge.net/datasets.html and LibSVM http://www.csie.ntu.edu.tw/~cjlin/libsvmtools/datasets/multilabel.html repositories.

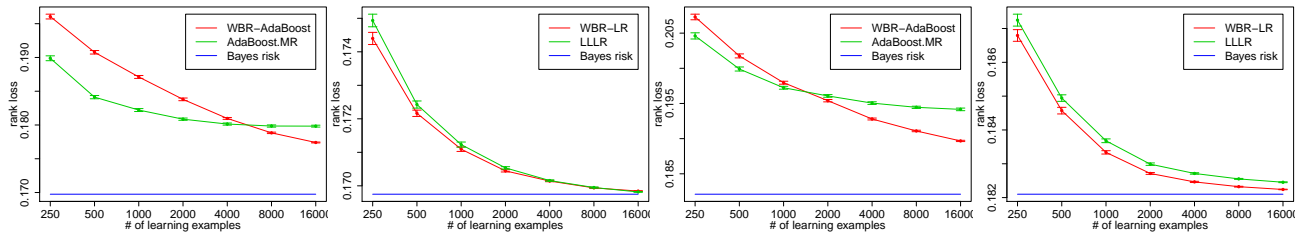

*Figure 1.* Learning curves on synthetic data. Two left plots present results on data with independent labels, while the two right plots on data with dependent labels. The plots compare the algorithms based on exponential loss separately from those based on logistic loss. The blue lines indicate the Bayes risk.

*Table 1.* Basic statistics for the datasets, including training and test set sizes, number of features and labels, and minimal, average, and maximal number of relevant labels.

| dataset | #train | #test | #attr. | #lab. | min | ave. | max |
| --- | --- | --- | --- | --- | --- | --- | --- |
| image | 1200 | 800 | 135 | 5 | 1 | 1.24 | 3 |
| emotions | 391 | 202 | 72 | 6 | 1 | 1.96 | 3 |
| scene | 1211 | 1196 | 294 | 6 | 1 | 1.06 | 3 |
| yeast | 1500 | 917 | 103 | 14 | 1 | 4.23 | 11 |
| mediamill | 30993 | 12914 | 120 | 101 | 0 | 4.36 | 18 |

*Table 2.* Experimental results on benchmark datasets in terms of rank loss. We compare the algorithms based on exponential loss (left) separately from those based on logistic loss (right). For each dataset, the winner of the two competing algorithms is marked by a *.

| dataset | AB.MR | WBR-AB | LLLR | WBR-LR |
| --- | --- | --- | --- | --- |
| image | 0.2081 | *0.2041 | *0.2047 | 0.2065 |
| emotions | 0.1703 | *0.1699 | 0.1743 | *0.1657 |
| scene | *0.0720 | 0.0792 | 0.0861 | *0.0793 |
| yeast | 0.2072 | *0.1820 | *0.1728 | 0.1736 |
| mediamill | 0.0665 | *0.0609 | 0.0614 | *0.0472 |

## 6. Conclusions

In this paper, we have shown that common univariate convex surrogates are consistent for mutlilabel ranking. We proved explicit regret bounds, relating ranking regret to univariate loss regret, which not only help to answer the question of consistency, but also inform about the rates of convergence.

For several reasons, our results should be of interest to the machine learning community. Most notably, because they are arguably surprising in light of (Gao & Zhou, 2011), where inconsistency is shown for the most popular pairwise surrogates. Moreover, on the more practical side, our results motivate simple and scalable algorithms for multilabel ranking, which are plain modifications of standard algorithms for classification (such as logistic regression or AdaBoost).

**Acknowledgments.** Krzysztof Dembczyński is supported by the Foundation of Polish Science under the Homing Plus programme, co-financed by the European Regional Development Fund. Wojciech Kotłowski is supported by the grant 91-517/DS funded by the Polish Ministry of Science and Higher Education. Eyke Hüllermeier is supported by German Research Foundation (DFG).